# From Coin to Data: The Impact of Object Detection on Digital Numismatics

Rafael Cabral[1], Maria De Iorio[1], Andrew Harris[1]

[1]Department of Paediatrics, Yong Loo Lin School of Medicine, National University of Singapore, Singapore

**Abstract**

In this work we investigate the application of advanced object detection techniques to digital numismatics, focussing on the analysis of historical coins. Leveraging models such as Contrastive Language–Image Pre-training (CLIP), we develop a flexible framework for identifying and classifying specific coin features using both image and textual descriptions. By examining two distinct datasets, modern Russian coins featuring intricate "Saint George and the Dragon" designs and degraded 1st millennium AD Southeast Asian coins bearing Hindu-Buddhist symbols, we evaluate the efficacy of different detection algorithms in search and classification tasks. Our results demonstrate the superior performance of larger CLIP models in detecting complex imagery, while traditional methods excel in identifying simple geometric patterns. Additionally, we propose a statistical calibration mechanism to enhance the reliability of similarity scores in low-quality datasets. This work highlights the transformative potential of integrating state-of-the-art object detection into digital numismatics, enabling more scalable, precise, and efficient analysis of historical artifacts. These advancements pave the way for new methodologies in cultural heritage research, artefact provenance studies, and the detection of forgeries.

## 1. Introduction

The use of computer vision methods in digital numismatics, the study of coinage, highlights the growing role of digital technology in the humanities over the past two decades (Schreibman, Siemens, and Unsworth, 2004; Berry, 2011). Defined by Drucker (2021) as the convergence of computational methods and humanities-focused research, "Digital Humanities" approaches have enabled researchers to address complex questions through large-scale data analysis and automation. Automated die studies in numismatics exemplify this, showcasing how computer vision and machine learning techniques can tackle the traditionally labour-intensive task of identifying coins struck from the same die. Historically, this process requires experts to manually examine coins for subtle similarities and differences, making it both time-consuming and subjective (Heinecke et al., 2021). Advances in image processing now allow for systematic analysis of die characteristics, such as engraver marks, symbols, and minor design inconsistencies, linking coins with greater precision and efficiency than ever before (McCord-Taylor 2020; Natarajan et al., 2023; Harris et al., 2024). By leveraging feature extraction algorithms and neural networks to compare and cluster coins from the same die, automated die studies have significantly improved both efficiency and consistency.

Building on these advancements, *image detection* methods are emerging as pivotal tools in digital numismatics, offering more efficient methods for identifying and classifying coins across large collections. Traditionally, numismatists rely on manual examination of physical coins or photographs which is a labour-intensive process complicated by the volume of coins, their varied conditions, and subtle design differences. Identifying details on worn coins or distinguishing intricate iconography has long been a challenge, particularly when working with extensive image databases. This reliance on manual methods not only slowed the categorization process but also limited studies to smaller datasets (Harris et al. 2024; Heinecke et al,. 2021). Automated die studies are achieving ever-increasing levels of accuracy (Cabral et al., 2024; Natarajan et al., 2023), driven by advancements in computational techniques and machine learning. However, the effectiveness of these studies can be significantly enhanced by integrating more robust and efficient object detection methods. Improved object detection not only facilitates the precise identification and classification of die features but also contributes to higher reliability and scalability of the analysis, particularly in complex or high-volume datasets. Incorporating state-of-the-art algorithms and addressing challenges such as occlusion, varying illumination, and die irregularities can further optimize the performance of automated systems in this domain. Moreover, image detection enables researchers to locate and classify coins based on distinct visual features, such as symbols, inscriptions, or stylistic elements, without requiring individual inspection. This greatly enhances efficiency, especially when searching for coins with specific motifs or iconography. Modern digital image detection, powered by Large Multimodal Models (advanced AI systems designed to process and integrate information from multiple data modalities, such as text, images, audio, and video; Goyal et al., 2023), allows users to query large databases using either reference images or descriptive text, such as "coin with Hindu script" or "symbol resembling a trident." This capability streamlines the organization and analysis of coins by mint, region, or historical period, facilitating the study of die links and regional coinage variations.

Furthermore, image detection tools support comparative studies across collections by identifying coins of similar origin or design. By incorporating statistical validation measures, these tools improve the accuracy of identifications, even for degraded or worn coins. This integration of advanced AI-driven image detection provides a more accessible, accurate, and efficient approach to numismatic research, transforming the field in the digital age.

In this paper, we address the challenge of detecting specific objects within images, with a focus on historical coins exhibiting significant degradation or erosion (Coin Set 1) and the detection of complex imagery, in this case "Saint George and the Dragon" (Coin Set 2). These scenarios have been chosen due to their limited exploration in the literature, which often prioritizes relatively modern coins featuring stylized decorations (although see Kampel et al., 2009; Huber-Mörk et al., 2012).

Object detection plays a crucial role in search tasks, especially when identifying images containing specific objects within large datasets. Manually reviewing each image in such cases is highly impractical. This capability is particularly valuable in fields such as numismatics, where it can assist in identifying coins with specific symbols, or in ceramic analysis, aiding in the identification of artefacts with similar decorative patterns. Such advancements contribute to a better understanding of trade routes and cultural exchanges. Additionally, object detection can assist in classifying coins into distinct groups, as coins from different dies, regions, or time periods are often characterised by unique decorative patterns. These tools are particularly useful in die studies and broader coin classification tasks, providing valuable insights into numismatic research.

As an alternative, we also explore a method that enables object searches based on textual descriptions. This approach involves describing complex shapes, symbols, or providing general attributes, such as "a small copper coin with an [x] shape and a [y] symbol on the right" or "a coin with Hindu iconography." This text-based search strategy has wide applicability, offering an intuitive way to locate specific objects in large datasets, especially when visual details are difficult to articulate or interpret.

## 2. Objectives

In this paper, we focus on two main tasks:

1. **Search Task:** Rank images based on their likelihood of containing the object of interest.
2. **Classification Task:** Classify coin images by determining the presence of the object of interest.

Both tasks are described in detail below. Our overarching goal is to develop a versatile framework for object detection within images, leveraging advanced techniques such as CLIP (Contrastive Language–Image Pre-training) models, alongside more traditional computer vision tools for object detection based on the SIFT, ORB, algorithms and the cosine distance. Specifically, we create a pipeline for object detection that operates using either reference images of the objects or textual descriptions of those objects.

*Similarity-Based Detection*

Initially, we calculate the similarity between the object of interest and sub-images extracted from the target images. We compare various models, pretrained datasets, and object types. However, challenges arise when noisy or low-quality images, suboptimal models, or vague object descriptions result in relatively low similarity scores, even when the object is present in the image. The difficulty lies in calibrating these similarity scores to determine whether an object is indeed present in the image. This calibration depends on factors such as image quality, model accuracy, and the specificity of object descriptions. For instance, a similarity score of 0.3 might be significant for low-quality images or coarse descriptions but insufficient for higher-quality images or more detailed descriptions. Generally, similarity measures are not directly interpretable and are primarily useful for ranking images, ensuring that the best matches appear at the top in search tasks.

*Calibration of Similarity Scores*

To address the challenge of interpreting similarity scores, we propose a p-value calculation algorithm to evaluate the "significance" of the similarity between objects and sub-images. This method involves comparing the observed similarity score with a distribution of "null" similarities, i.e., the distribution of scores under the hypothesis that the object is absent. Our approach is analogous to a permutation test: by generating null similarity distributions, we can calculate p-values to assess whether the observed similarity score is statistically significant. This allows us to determine the presence of an object in an image by applying a simple threshold to the p-values.

Advantages of the Proposed Method include (i) flexibility, as the pipeline allows for both reference images and textual descriptions of objects to be detected; (ii) robustness, as the statistical p-value provides a more reliable criterion for object presence, especially in noisy or low-quality data and (iii) scalability, since the ranking system ensures efficient image retrieval in search tasks, even with large datasets. By integrating advanced object detection techniques and statistical calibration, our framework represents a significant step forward in handling complex object detection tasks, particularly in challenging scenarios such as historical coin analysis.

## 3. Coin Datasets

The two sets of coin images examined in this study differ greatly in both time and production technique, which helps illustrate the efficacy of the methods discussed. The first set includes images of 608 late 20$^{th}$ and 21$^{st}$ century machine minted small denomination coins (such as cents, pesos, paise, etc.) from various countries (Wanderdust, n.d.). The image of St. George slaying a Dragon, on horseback, is found within this set only on a limited series of limited series of Russian 50, 10, 5, and 1 kopeck coins issued in the early 2000s. The Cyrillic letters М (Moscow) or С-П (Saint Petersburg) positioned below the horse's front hoof, indicating the mint of origin. The inscription "БАНК РОССИИ" (BANK OF RUSSIA) runs along the rim, divided by the figure of St. George, while the year of mintage is displayed beneath the serpent (Numista, n.d.). The St. George motif is found on Moscow's coat of arms and is symbolic of courage, protection, victory, and religious reverence in Russian Orthodox Christianity. While the image of St. George is typically associated with higher-value Russian coinage, its appearance on these low-denomination coins has elevated them to collector's items, also linked to superstitions regarding their status as good luck charms in addition to their symbolism of national heritage and continuity.

The second set comprises images of 521 die-struck silver coins from early 1st millennium AD Southeast Asia, found online in auction databases (ACSearch, n.d.). These coins feature images of *svastika*/swastika, a prominent Hindu-Buddhist symbol associated with cosmic order, spiritual protection, and royal legitimacy, a motif represents the cyclical nature of the universe and the eternal processes of creation, preservation, and destruction (Gutman 1978, Wicks 1992)[1]. Single swastika are depicted on the reverse of these coins alongside *bhadrapittha* (auspicious seat), flanking a larger Srivatsa (aniconic Indic symbol), with the obverse typically showcasing a conch or half-risen sun surrounded by beads inspired by Vedic astrology. Swastika-bearing coins, initially minted by the Pyu people of north-central Myanmar, circulated extensively throughout Myanmar, Thailand, Cambodia, and Vietnam (Epinal & Gardère, 2014), and became a prominent form of currency across Southeast Asia during a time of significant, though poorly documented, maritime trade and prosperity (Miksic & Goh, 2017). These coins were instrumental in fostering economic and cultural exchanges across the region, reflecting the broader spread of Indic culture and political-religious influence, also known as Indianisation (Cœdès, 1968).

---

[1] The authors would like to sincerely emphasize the difference between the *svastika/*swastika, used in this context by Hindu and Buddhist populations around the world for millennia as a sacred symbol, and the 20th century *Hakenkreuz*, a problematic and rightly condemned symbol of hate and historical atrocity.

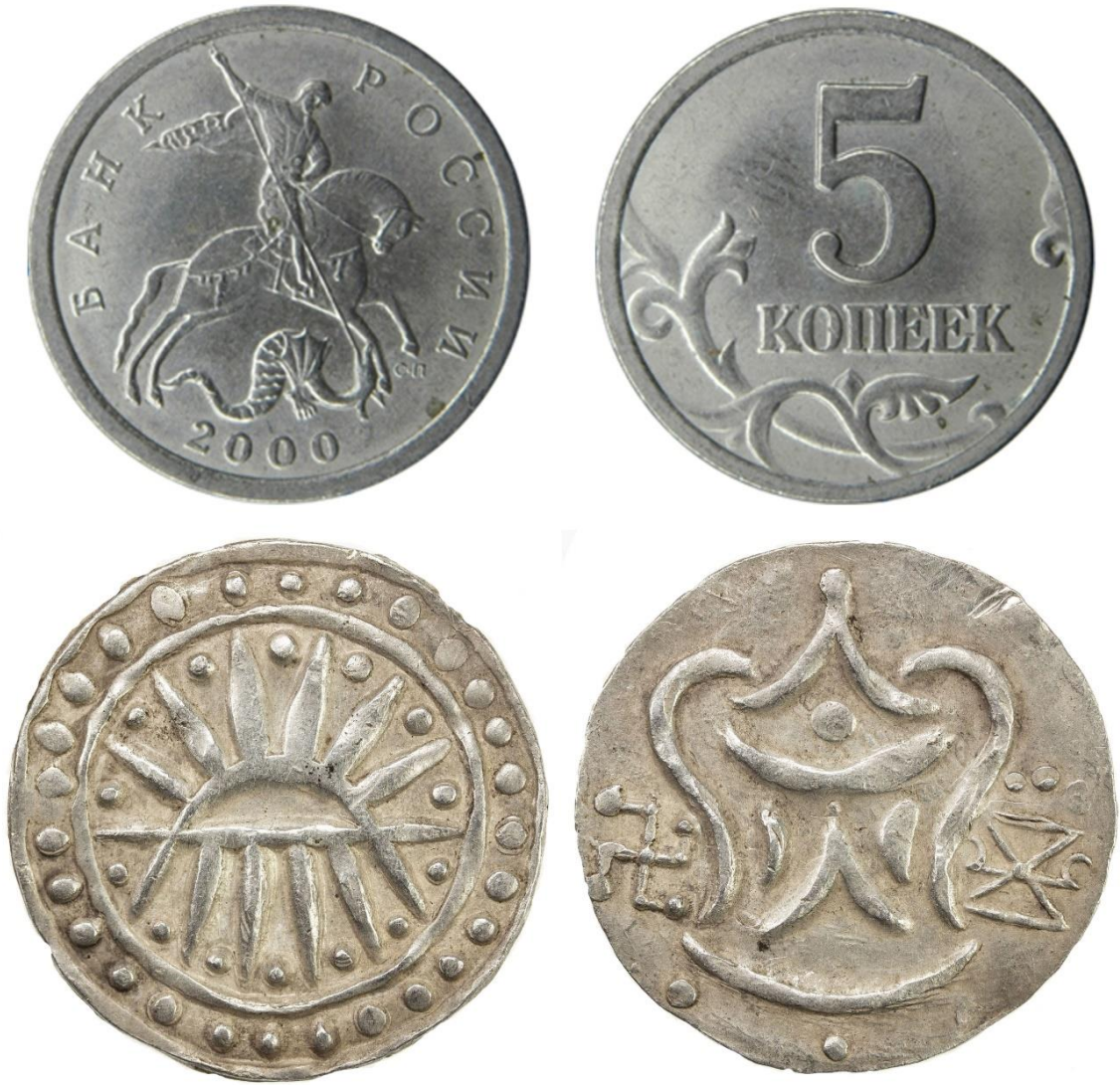

**Figure 1**
Above: Example 5 Kopeck Coin (2000)
(https://en.numista.com/catalogue/pieces1218.html)
Below: Example Rising Sun/*Srivatsa* coin, Myanmar (1st millennium AD) (Hindu-Buddhist *svastika/swastika* image centre-left) (https://www.sarc.auction/HALIN-late-5th-early-6th-century-AR-unit-9-58g_i40779664)

## 4. Algorithms and Techniques

We compare different models for object detection, using both an object and/or a text description as input. The models are summarised in Error! Reference source not found..

Contrastive Language-Image Pre-training (CLIP) is a Large Multimodal model jointly developed by OpenAI and UC Berkeley that bridges the gap between text and image modalities (see Radford et al., 2021). By embedding text and images into a shared space using a contrastive learning objective, CLIP enables a wide range of applications across domains such as cross-modal retrieval, text-to-image generation, image captioning, aesthetic ranking, and zero-shot image classification. Moreover, CLIP has become the backbone for multimodal large language models (LMMs), where its ability to align vision and language inputs enhances language models with visual understanding.

The core idea behind CLIP is training a pair of neural networks, a text encoder and an image encoder, on large-scale datasets of image-text pairs. The model minimises the distance between embeddings of matching image-text pairs while maximising the distance for non-matching pairs, effectively learning to align visual and textual data. This approach allows CLIP to excel in zero-shot learning, enabling tasks like image classification or retrieval without explicit task-specific training. Unlike traditional captioning models, CLIP does not generate captions but instead determines how well a given description fits an image.

CLIP's training method draws from contrastive self-supervised learning, which leverages positive and negative examples. For instance, in a binary classification task to identify birds, positive examples might include images of birds, while negative examples would be images without birds. This strategy ensures robust learning of visual concepts directly from natural language supervision.

In this work, we employ the open source implementations in [GitHub - mlfoundations/open_clip: An open source implementation of CLIP.](#) ( [[2212.07143] Reproducible scaling laws for contrastive language-image learning (arxiv.org)](#) . See [open_clip/docs/openclip_results.csv at main · mlfoundations/open_clip · GitHub](#) for more details about the models implemented. The words "large", "medium", "small" in table are relative to the models available in the "open_clip" distribution.

We also evaluate two image matching techniques: **Scale-Invariant Feature Transform (SIFT)** and **Oriented FAST and Rotated BRIEF (ORB)**. Both algorithms are widely used for detecting and describing keypoints in images, but they employ distinct methods and have different strengths.

SIFT (Lowe 2004) is one of the most renowned feature detection and description algorithms. It identifies keypoints using the **Difference of Gaussians (DoG) methods**, an approximation of the Laplacian of Gaussian (LoG). Keypoints are refined through Taylor series expansion, with orientations assigned based on local gradients to ensure robustness against scale and rotation. Descriptors are computed by dividing a 16x16 neighbourhood into 4x4 sub-regions, creating a 128-dimensional vector. The similarity score between images is then obtained by computing the inverse of the Euclidian distance between the 128-dimensional vectors. Although highly accurate and robust, SIFT is computationally intensive, limiting its use in real-time applications.

ORB (Rublee et al. 2011) combines the **FAST** (Features from Accelerated Segment Test) keypoint detector with the **BRIEF** (Binary Robust Independent Elementary Features) descriptor. Keypoints are ranked using the **Harris corner score**, and rotational invariance is achieved by aligning BRIEF descriptors to keypoint orientations. ORB produces binary descriptors that are computationally efficient, making it well-suited for real-time tasks. It is robust to scale, rotation, and limited affine transformations.

While SIFT can offer superior accuracy and robustness to a wide range of transformations, its computational intensity is a significant drawback. ORB, with its efficient binary descriptors, provides a faster and more practical solution for applications requiring real-time performance. Finally, we consider the cosine similarity between the target (sub)images and the object image.

| object descriptor type | model_name | Description |
| --- | --- | --- |
| Images | ViT-H-14-378-quickgelu | "Large" CLIP model with 986.71M parameters trained on "large" training data "DFN-5B". We compare the embeddings of (sub)images with the embedding of the image of the object |
| Text | ViT-H-14-378-quickgelu | Same as before, but we compare the embeddings of (sub)images with the embedding of the text of the object |
| Text | ViT-L-14-quickgelu | "Medium" CLIP model with 427.62M parameters trained on "medium" training data "DFN-2B". We compare the embeddings of (sub)images with the embedding of the image of the object |
| Images | ViT-L-14-quickgelu | Same as before, but we compare the embeddings of (sub)images with the embedding of the text of the object |
| Images | ViT-B-32 | "Small" CLIP model with 151.28M parameters trained on "small" training data "laion2b_s34b_b79k". We compare the embeddings of (sub)images with the embedding of the image of the object |
| Text | ViT-B-32 | Same as before, but we compare the embeddings of (sub)images with the embedding of the text of the object |
| Image | Orb | We compute "ORB" keypoints for the target (sub)images and objects and then compute the number of matches between keypoints: more matches -> more similarity |
| Image | SIFT | We compute "SIFT" keypoints for the target (sub)images and objects and then compute the number of |

|  |  | matches between keypoints: more matches -> more similarity |
| --- | --- | --- |
| Image | Cosine | We compute the cosine similarity directly between the target (sub)images and the object image |

Table 1: Implemented Models

## 5. Results

The workflow which attained these results can be found in the **Appendix** accompanying this paper. We have developed this entire workflow as a Python package with user-friendly functions for seamless implementation (https://drive.google.com/drive/folders/1Y_rVuaboLDeqiHJ0seVzDwOQ6v9L0aKL). The CLIP model functionality is built using the open clip library, while the SIFT and ORB approaches leverage the cv2 library, which is part of the widely used OpenCV framework (OpenCV-python package available on PyPI).

### 5.1 Evaluation Metrics

The reference objects we used for the Russian and Southeast Asian coin sets are presented in **Section 3** of the **Appendix** for this paper. To evaluate the performance of the search task, we measure the number of coins containing the target objects among the top-ranked coins based on similarity. Specifically, for coin set 1, we assess the top 20 coins, while for coin set 2, we focus on the top 10 coins. For the classification task, we use P-value thresholds of 0.01, 0.05, and 0.1 to determine whether the object is present on a coin. The classification results are then compared against the ground truth, and we compute metrics such as specificity, sensitivity, balanced accuracy, and the F1-score, with a preference for the F1-score due to its balanced consideration of precision and recall.

### 5.2 Findings for Coin Set 1

We begin by examining the "matches in first 10" column in Table 3, which evaluates the effectiveness of object detection procedures in search tasks. This metric is critical because, in search tasks, we aim for the best matches to appear at the top of the ranked list based on similarity to the object of interest.

In **Table 2**, the "large" (ViT-H-14-378-quickgelu) and "medium" (ViT-L-14-quickgelu) CLIP models consistently identified all 8 correct coins within the top 10, regardless of whether the object was represented as an "image" or "text." In contrast, the "small" CLIP model (ViT-B-32) identified only 4 out of the 8 coins, while other approaches failed to detect any of the correct coins. These results highlight the superior performance of the larger CLIP models in this context.

| object type | model_name | matches in first 10 |
| --- | --- | --- |
| images | ViT-H-14-378-quickgelu | 8 |

| | ViT-H-14-378- | |
|---|---|---|
| text | quickgelu | 8 |
| text | ViT-L-14-quickgelu | 8 |
| images | ViT-L-14-quickgelu | 8 |
| images | ViT-B-32 | 4 |
| text | ViT-B-32 | 4 |
| Orb | | 1 |
| SIFT | | 0 |
| Cosine | | 0 |

**Table 2: Evaluation metrics for the different object detection approaches for coin set 1**

Then, we classify a coin as containing the object if the derived P-value is smaller than a specified threshold. Among the thresholds tested (0.01, 0.05, and 0.1), the best overall F1-score was achieved with a threshold of 0.01. The model that performed best in terms of both F1-score and balanced accuracy was the ViT-L-14-quickgelu model using the third P-value calculation mechanism (see **Table 3**) The top five results were consistently obtained by using either the "medium" or "large" CLIP models (ViT-L-14-quickgelu or ViT-H-14-378-quickgelu), in combination with either the first or third P-value calculation mechanism, and employing either image or text formats for the object representations. In contrast, all other models and P-value mechanisms exhibited significantly poorer performance.

| object type | model_name | Pvalue type | matches in first 10 | balanced_accuracy001 | f1_score001 |
|---|---|---|---|---|---|
| text | ViT-L-14-quickgelu | 3 | 8 | 0.999167 | 0.941176 |
| images | ViT-L-14-quickgelu | 3 | 8 | 0.998333 | 0.888889 |
| text | ViT-L-14-quickgelu | 1 | 8 | 0.9975 | 0.842105 |
| images | ViT-H-14-378-quickgelu | 3 | 8 | 0.996667 | 0.8 |
| images | ViT-H-14-378-quickgelu | 1 | 8 | 0.9925 | 0.64 |
| images | ViT-L-14-quickgelu | 2 | 8 | 0.6875 | 0.545455 |
| images | ViT-B-32 | 1 | 4 | 0.685833 | 0.461538 |
| text | ViT-B-32 | 1 | 4 | 0.745 | 0.444444 |
| images | ViT-B-32 | 2 | 4 | 0.863333 | 0.428571 |
| text | ViT-B-32 | 2 | 4 | 0.921667 | 0.411765 |
| text | ViT-B-32 | 3 | 4 | 0.920833 | 0.4 |
| text | ViT-L-14-quickgelu | 2 | 8 | 0.974167 | 0.340426 |
| images | ViT-B-32 | 3 | 4 | 0.74 | 0.333333 |
| text | ViT-H-14-378-quickgelu | 1 | 8 | 0.965 | 0.275862 |
| text | ViT-H-14-378-quickgelu | 3 | 8 | 0.946667 | 0.2 |
| images | ViT-H-14-378-quickgelu | 2 | 8 | 0.941667 | 0.186047 |

| | ViT-H-14-378- | | | | |
|---|---|---|---|---|---|
| text | quickgelu | 2 | 8 | 0.9375 | 0.175824 |
| Orb | | 3 | 1 | 0.498333 | 0 |
| Cosine | | 3 | 0 | 0.494167 | 0 |
| images | ViT-L-14-quickgelu | 1 | 8 | 0.5 | 0 |
| SIFT | | 1 | 0 | 0.490833 | 0 |
| Orb | | 1 | 1 | 0.5 | 0 |
| Cosine | | 1 | 0 | 0.5 | 0 |
| SIFT | | 3 | 0 | 0.498333 | 0 |

**Table 3: Number of coins with a visible swastika in the top 20 entries for the different object detection approaches.**

In **Figure 2** we present the computed similaritie for three groups: the reference coins (blue), the coins in the dataset that do not contain the "Saint George and the Dragon" scene, and the coins that do contain it (orange). These results correspond to the best-performing model and P-value calculation mechanism based on the F1-score.

The figure illustrates an ideal scenario where the similarities for the correct coins (those containing the object or symbol) are distinctly higher than both the reference distribution similarities and the similarities of unmatched coins. For a search task to be effective, the similarities of the correct coins must be higher than those of the incorrect coins to ensure that the correct coins appear first in the ranked results. Additionally, the similarity distribution for the incorrect coins should align closely with the reference distribution, ensuring that low P-values are assigned exclusively to the correct coins, thus indicating statistical significance.

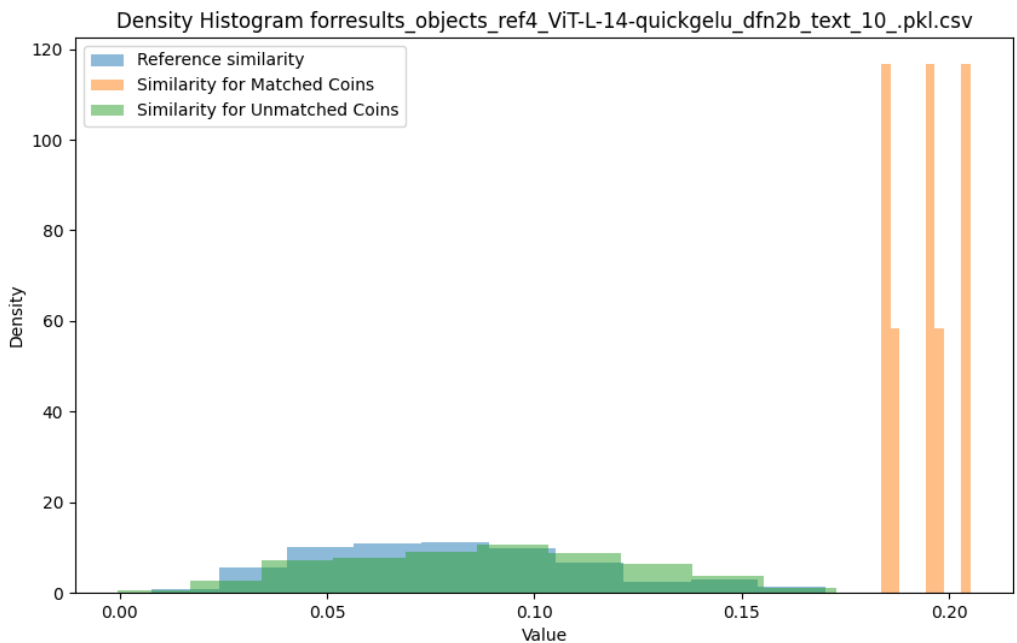

**Figure 2: Histograms of the similarities for different coin groups for the Vit-L-14-quickgelu model with "text" objects with p-value calculation mechanism 3.**

In Figure 3, we present the similarity distributions for models and P-value mechanisms that failed to perform effectively. In the first scenario, the similarities for the correct coins are lower than those of the reference distribution. While the correct coins may occasionally appear at the top in search tasks, the P-values derived from these similarities are not meaningful for classification purposes. In the second scenario, there is minimal differentiation between the distributions of correct coins, incorrect coins, and the reference distribution, making it impossible to reliably distinguish the correct coins based on similarity. These results highlight the importance of both high similarity values for correct coins and clear separation between distributions for successful detection and classification.

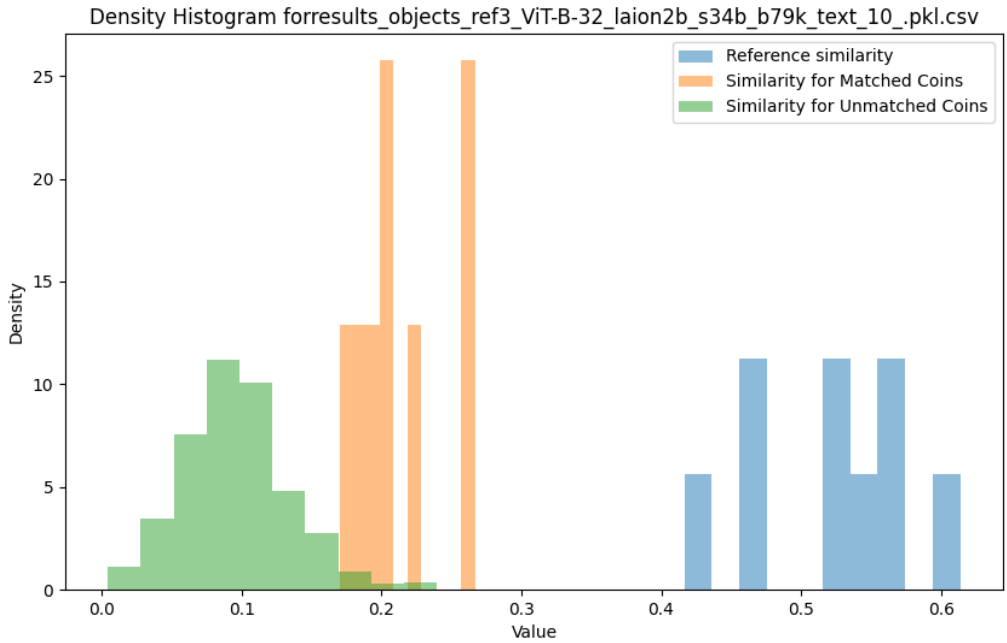

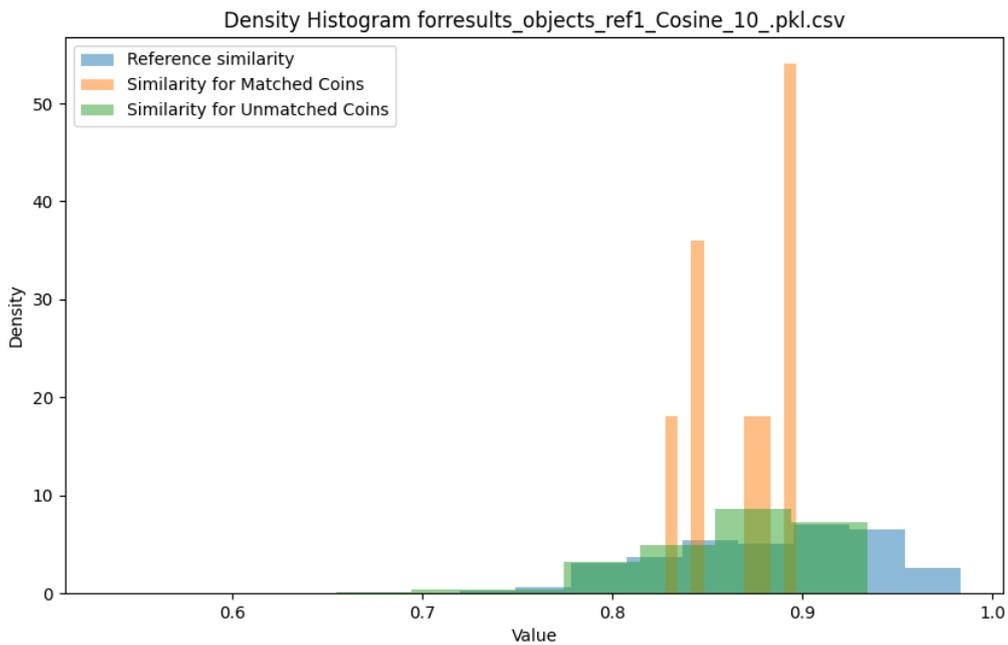

**Figure 3: Histograms of the similarities for different coin groups for the Vit-B-32 model with "text" objects with p-value calculation mechanism 3 (top) and similarities computed via the cosine similarity with p-value mechanism 3 (bottom).**

In Figure 4 we summarise the output file for the best performing model according to the f1-score.

| Image | Name | Text | Similarity | p-value |
|---|---|---|---|---|
| 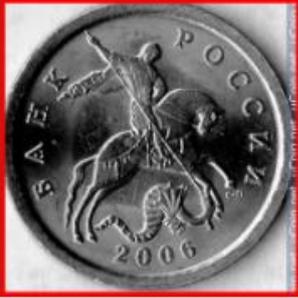 | 035__5 Kopeks_russia.jpg | Saint George and the Dragon | 0.2052 | 0.0000 |
| 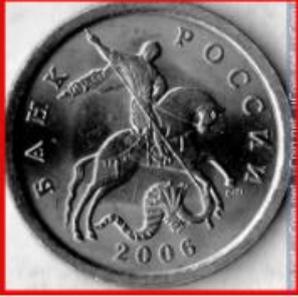 | 003__5 Kopeks_russia.jpg | Saint George and the Dragon | 0.2052 | 0.0000 |
| 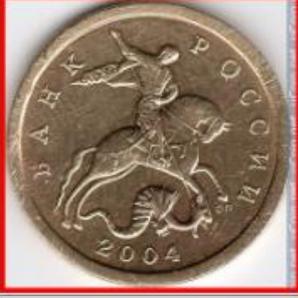 | 016__10 Kopeks_russia.jpg | Saint George and the Dragon | 0.1983 | 0.0000 |
| 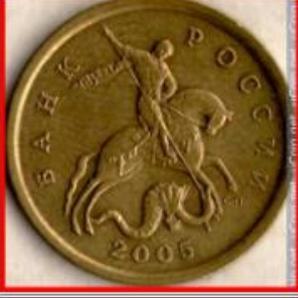 | 002__50 Kopeks_russia.jpg | Saint George and the Dragon | 0.1945 | 0.0000 |
| 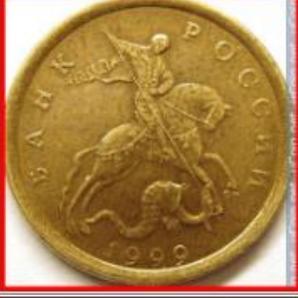 | 025__10 Kopeks_russia.jpg | Saint George and the Dragon | 0.1945 | 0.0000 |

**Figure 4: Top 6 coins with highest similarity for coin set 1 with respective p-values. Image source: https://www.kaggle.com/datasets/wanderdust/coin-images**

## 5.3 Findings for Southeast Asian coin set

We begin by examining the "matches in first 20" column in **Table 4**. In the subsequent table, the "small" CLIP model (ViT-B-32) using object representations in the "text" format achieved the best performance. This was followed by the "medium" CLIP model (ViT-L-14-quickgelu) utilizing the swastika object in the "image" format. These results highlight the effectiveness of both text-based and image-based approaches, with the smaller model excelling in textual representation and the medium-sized model performing well with visual input.

| object type | model_name | matches in first 20 |
|---|---|---|
| text | ViT-B-32 | 20 |
| images | ViT-L-14-quickgelu | 15 |
| images | ViT-H-14-378-quickgelu | 13 |
| images | ViT-B-32 | 11 |
| Cosine | | 11 |
| text | ViT-H-14-378-quickgelu | 11 |
| text | ViT-L-14-quickgelu | 11 |
| Orb | | 9 |
| SIFT | | 5 |

**Table 4: Evaluation metrics for the different object detection approaches for coin set 2.**

Then, we classify a coin as containing the object if the derived P-value is smaller than a specified threshold. Among the thresholds tested (0.1, 0.05, and 0.01), the threshold of 0.1 yielded the best overall F1-score (see **Table 5**). The best-performing model in terms of both F1-score and balanced accuracy was the ViT-L-14-quickgelu model using the third P-value calculation mechanism. The top five results were achieved using either the "medium" or "large" CLIP models (ViT-L-14-quickgelu or ViT-H-14-378-quickgelu) with the first or third P-value mechanism, employing either images or text representations for the object. In contrast, all other models and P-value mechanisms showed significantly poorer performance.

The "ORB," "Cosine," and "SIFT" methods performed better for the Russian second coin, likely due to the simplicity of the swastika, a geometric shape that is more easily detected using direct methods. This contrasts with the "Saint George and the Dragon" scene, which is much more complex and requires more sophisticated models to achieve accurate detection.

| object type | model_name | Pvalue type | matches in first 20 | balanced_accuracy010 | f1_score010 |
|---|---|---|---|---|---|
| Text | ViT-H-14-378-quickgelu | 2 | 11 | 0.575288 | 0.685294 |
| images | ViT-L-14-quickgelu | 2 | 15 | 0.523061 | 0.680054 |
| Text | ViT-B-32 | 2 | 15 | 0.633212 | 0.677029 |
| Text | ViT-L-14-quickgelu | 2 | 11 | 0.606364 | 0.671053 |
| images | ViT-H-14-378-quickgelu | 2 | 13 | 0.5355 | 0.663793 |
| images | ViT-L-14-quickgelu | 1 | 15 | 0.525439 | 0.591216 |
| images | ViT-H-14-378-quickgelu | 1 | 13 | 0.537242 | 0.571429 |
| Text | ViT-H-14-378-quickgelu | 1 | 11 | 0.568424 | 0.534447 |
| Text | ViT-L-14-quickgelu | 3 | 11 | 0.5615 | 0.464789 |
| Text | ViT-B-32 | 3 | 15 | 0.589621 | 0.423592 |
| Text | ViT-L-14-quickgelu | 1 | 11 | 0.547409 | 0.407035 |
| images | ViT-L-14-quickgelu | 3 | 15 | 0.518894 | 0.348052 |
| Text | ViT-B-32 | 1 | 20 | 0.2 | 0.333333 |
| images | ViT-H-14-378-quickgelu | 3 | 13 | 0.525742 | 0.322404 |
| Text | ViT-H-14-378-quickgelu | 3 | 11 | 0.522485 | 0.293785 |
| Cosine | | 3 | 11 | 0.520591 | 0.288952 |
| SIFT | | 1 | 5 | 0.477333 | 0.243767 |
| images | ViT-B-32 | 2 | 11 | 0.500606 | 0.196319 |
| Orb | | 3 | 9 | 0.503136 | 0.171429 |
| images | ViT-B-32 | 3 | 11 | 0.495667 | 0.142395 |
| SIFT | | 3 | 5 | 0.476409 | 0.098361 |
| images | ViT-B-32 | 1 | 11 | 0.501045 | 0.064057 |
| Orb | | 1 | 9 | 0.503576 | 0.02963 |
| Cosine | | 1 | 11 | 0.5 | 0 |

**Table 5: Evaluation metrics for the different object detection approaches for coin set 2**

**Figure 5** illustrates the computed similarities for three groups: the reference coins (blue), coins in the dataset that do not contain a swastika (green), and coins that do contain the scene (orange). These results are based on the best-performing model according to the F1-score. The figure shows that the similarities for the correct coins (those containing the object/symbol) tend to be slightly larger than both the reference distribution similarities and the similarities for the incorrect coins. The figure reveals that the similarity distribution for unmatched coins is larger than the reference distribution, leading to low P-values for incorrect coins and, consequently, poorer algorithm performance in significance tasks.

Once again, these results suggest that while CLIP models perform well for search tasks, they are less effective for significance-based tasks in this coin set, likely due to the degraded condition of many coins, which makes accurate detection more challenging.

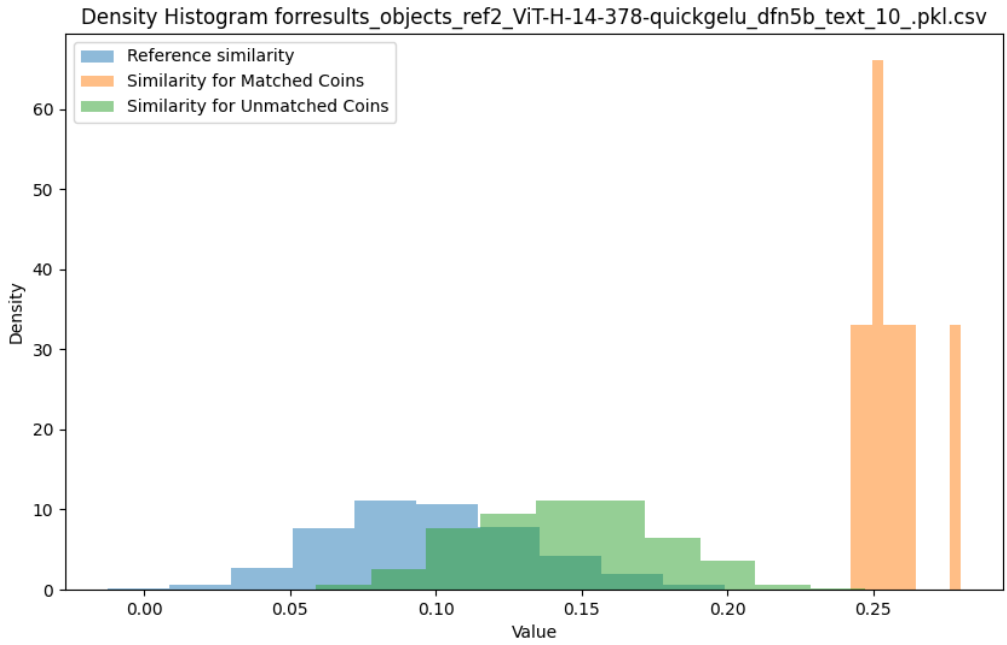

**Figure 5: Histograms of the similarities for different coin groups for the ViT-H-14-378-quickgelu models with "text" objects with p-value calculation mechanism 2.**

In **Figure 6** we summarise the output or the best performing model according to the f1-score:

| Image | Name | Text | Similarity | p-value |
|---|---|---|---|---|
| 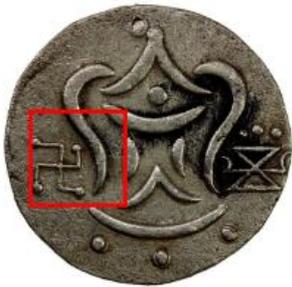 | RS0916 Reverse.jpg | swastika stamped on aged coinage | 0.4827 | 0.0000 |
| 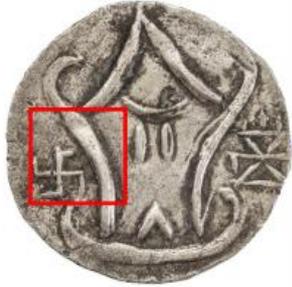 | RS0974 Reverse.jpg | swastika stamped on aged coinage | 0.4762 | 0.0000 |
| 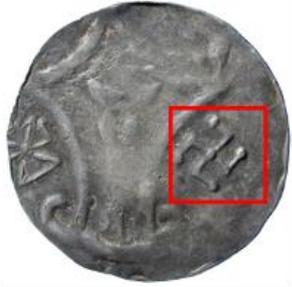 | RS0568 Reverse.jpg | swastika stamped on aged coinage | 0.4639 | 0.0000 |
| 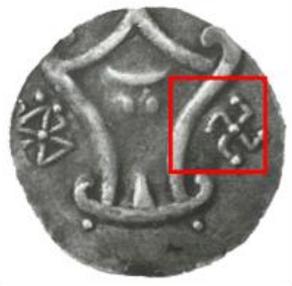 | RS0145 (Reverse).JPG | swastika stamped on aged coinage | 0.4566 | 0.0000 |
| 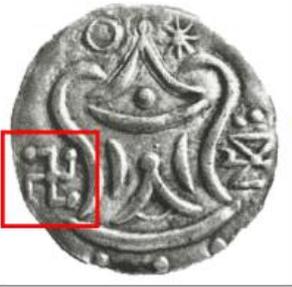 | RS0130 (Reverse).jpg | swastika stamped on aged coinage | 0.4564 | 0.0000 |
| 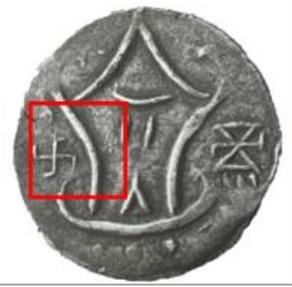 | RS0144 (Reverse).JPG | Eroded swastika symbol on an ancient coin | 0.4525 | 0.0000 |

**Figure 6: Top 6 coins with highest similarity for coin set 2 with respective p-values. URL: https://db.stevealbum.com/php/home.php?site=0&lang=1&cust=0**

## 6. Conclusion

In this work, we evaluate the performance of CLIP models and other traditional computer vision techniques for object detection tasks across two distinct coin sets. The results consistently demonstrated that CLIP models outperform traditional methods in all tasks. CLIP models prove particularly effective in search tasks, with larger models trained on extensive datasets yielding the best results. For both coin sets, the use of CLIP leads to desirable outcomes, underscoring their utility for such applications.

In classification tasks, CLIP models also perform very well for the Russian coin set, with performance depending on the specific P-value calculation mechanism employed. Traditional computer vision techniques, however, perform better on the Southeast Asian coin set, where the target object (the swastika) is a simple geometric shape. Conversely, these traditional methods fail entirely on the Russian coin set, where the target object (Saint George and the Dragon) is significantly more complex.

The classification results for Southeast Asian coin set are overall less satisfactory, primarily due to the degraded condition of many coins, which hindered accurate detection (**Figure 7**). To address such challenges, we propose and implement various empirical methods to assess the significance of object detection results, providing a framework for robust evaluation in similar tasks. This work highlights the strengths and limitations of both modern and traditional approaches, emphasising the importance of selecting appropriate models and evaluation mechanisms for different types of objects and datasets.

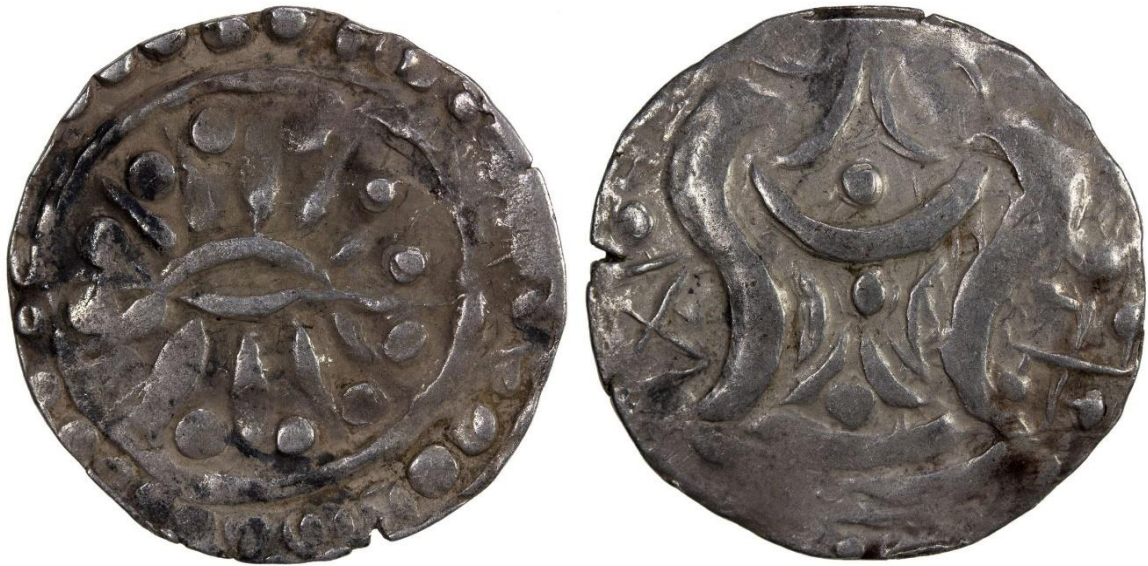

**Figure 7: Example of degraded Rising Sun/*Srivatsa* coin. *Svastika/*swastika only partially visible on right side. URL: https://www.sarc.auction/BEIKTHANO-9th-10th-century-AR-unit-8-67g_i34099681**

The continued development of these methods is poised to revolutionise the analysis of historical artefacts, particularly in digital numismatics where this approach not only improves coin classification and provenance studies but also facilitates broader investigations into historical trade networks, cultural exchanges, and economic systems through digitized, scalable methodologies (A. Harris et al., 2024), as well as detection of forgeries among existing collections of coins (Li et al., 2017). By overcoming the limitations of traditional computer vision in complex recognition tasks, advanced models enable precise and scalable artifact analysis. This integration not only enhances the identification and classification of coins but also supports broader cultural and historical inquiries, enriching our understanding of socio-economic networks and iconographic traditions. As digital methodologies evolve, combining machine learning with humanities research promises to unlock new dimensions of historical knowledge, preserving and reinvigorating the study of human history through interdisciplinary, data-driven approaches.

## Bibliography

Berry, D. M. (2011). The computational turn: Thinking about the digital humanities. *Culture Machine, 12*, 1–22.

Cabral, R., De Iorio, M., & Harris, A. (2024). *Scalable Bayesian clustering for integrative analysis of multi-view data*. arXiv. https://doi.org/10.48550/arXiv.2408.17153

Chen, H.-Y., Lai, Z., Zhang, H., Wang, X., Eichner, M., You, K., Cao, M., Zhang, B., Yang, Y., & Gan, Z. (2024). *Contrastive localized language-image pre-training* [Preprint]. arXiv. https://doi.org/10.48550/arXiv.2410.02746

Cœdès, G. (1968). *The Indianized states of Southeast Asia* (S. B. Cowing, Trans.). University of Hawaii Press.

Drucker, J. (2021). *Digital humanities coursebook: An introduction to digital methods for research and scholarship*. Oxford University Press.

Epinal, G., & Gardère, J. D. (2014). Cambodia from Funan to Chenla: A thousand years of monetary history (pp. 93–125). Phnom Penh: SOSORO Museum of Money and Economy.

Goyal, S., Kiela, D., & Fu, D. (2023). *Scaling Multimodal Models: Insights and Innovations*. Trends in Artificial Intelligence.

Gutman, P. (1978) The Ancient Coinage of Southeast Asia. *JSS* 066(1c), 8-21

Harris, A., Cremaschi, A., Lim, T.S., De Iorio, M., & Kwa, C.G. (2024). From past to future: Digital methods towards artefact analysis. *Digital Scholarship in the Humanities, 39*(4), 1026–1042. https://doi.org/10.1093/llc/fqae057Heinecke et al. 2021

Huber-Mörk, R., Nlle, M., Rubik, M., Hdlmoser, M., Kampel, M., & Zambanini, S. (2012). Automatic Coin Classification and Identification. InTech. doi: 10.5772/35795

Kampel M., Huber-Mörk, R. and Zaharieva, M. (2009). "Image-Based Retrieval and Identification of Ancient Coins." *IEEE Intelligent Systems*, 24(2), 26-34, DOI: 10.1109/MIS.2009.29

Liu, L., Lu, Y., & Suen, C. Y. (2017). An image-based approach to detection of fake coins. *IEEE Transactions on Information Forensics and Security, 12*(5), 1227–1239. https://doi.org/10.1109/TIFS.2017.2656478

Miksic, J. N., & Goh, G.Y. (2017). *Ancient Southeast Asia*. Routledge.

Natarajan, A., De Iorio, M., Heinecke, A., Mayer, E., & Glenn, S. (2023). Cohesion and repulsion in Bayesian distance clustering. *Journal of the American Statistical Association*, *118*(543), 1374–1384. https://doi.org/10.1080/01621459.2023.2191821

Numista (n.d.) 5 Kopecks. Retried December 5, 2024 from https://en.numista.com/catalogue/pieces1218.html

Radford, A., Kim, J. W., Hallacy, C., Ramesh, A., Goh, G., Agarwal, S., Sastry, G., Askell, A., Mishkin, P., Clark, J., Krueger, G., & Sutskever, I. (2021). Learning transferable visual models from natural language supervision. *Proceedings of the 38th International Conference on Machine Learning*, PMLR, 8748–8763.

Rublee, E., Rabaud, V., Konolige, K., & Bradski, G. (2011). ORB: an efficient alternative to SIFT or SURF. *IEEE International Conference on Computer Vision (ICCV 2011)*, 2564–2571. https://doi.org/10.1109/ICCV.2011.6126542:contentReference[oaicite:0]{index=0}:contentReference[oaicite:1]{index=1}.

Schriebman, S., Siemens, R., & Unsworth, J. (2004). *A companion to digital humanities*. Blackwell Publishing.

Stephen Album Rare Coins. (n.d.). *Stephen Album Rare Coins database*. Retrieved December 24, 2024, from https://db.stevealbum.com/php/home.php?site=0&lang=1&cust=0

Taylor, Z.M. (2020). *The Computer-Aided Die Study (CADS): A tool for conducting numismatic die studies with computer vision and hierarchical clustering* (Master's thesis, Trinity University). Trinity University Digital Commons. https://digitalcommons.trinity.edu/

Wanderdust. (n.d.). *Coin images* [Dataset]. Kaggle. https://www.kaggle.com/datasets/wanderdust/coin-images

Wicks, R. S. (1992). *Money, markets, and trade in early Southeast Asia: The development of indigenous monetary systems to AD 1400*. SEAP Publications. https://doi.org/ISBN 9780877277101